\documentclass[lettersize,journal]{IEEEtran}
\ifCLASSOPTIONcompsoc
  \usepackage[nocompress]{cite}
\else
  \usepackage{cite}
\fi
\ifCLASSINFOpdf
\else
\fi
\usepackage{amsmath}
\usepackage{amsfonts}
\usepackage{graphicx}
\usepackage{booktabs}
\usepackage[table]{xcolor}
\usepackage{makecell}

\usepackage{ragged2e}
\usepackage[colorlinks,linkcolor=red]{hyperref}

\begin{document}
%
\title{SimVPv2: Towards Simple yet Powerful Spatiotemporal Predictive Learning}
%
%
%
%

\author{Cheng~Tan$^*$,
        Zhangyang~Gao$^*$,
        Siyuan Li$^*$,
        and~Stan~Z.~Li,~\IEEEmembership{Fellow,~IEEE}
\IEEEcompsocitemizethanks{
\IEEEcompsocthanksitem * Equal contribution. \protect\\
\vspace{-2mm}
\IEEEcompsocthanksitem Cheng Tan, Zhangyang Gao and Siyuan Li are with Zhejiang University, Hangzhou, China, and also with the AI Lab, Research Center for Industries of the Future, Westlake University. Email: \{tancheng, gaozhangyang, lisiyuan\}@westlake.edu.cn.
\IEEEcompsocthanksitem Stan Z. Li is with the AI Lab, Research Center for Industries of the Future, Westlake University. Email: Stan.ZQ.Li@westlake.edu.cn. \protect\\
}
}

%
%

\markboth{Journal of \LaTeX\ Class Files,~Vol.~14, No.~8, July~2022}%
{Shell \MakeLowercase{\textit{et al.}}: Bare Demo of IEEEtran.cls for Computer Society Journals}
%


\maketitle

\begin{abstract}
Recent years have witnessed remarkable advances in spatiotemporal predictive learning, with methods incorporating auxiliary inputs, complex neural architectures, and sophisticated training strategies. While SimVP has introduced a simpler, CNN-based baseline for this task, it still relies on heavy Unet-like architectures for spatial and temporal modeling, which still suffers from high complexity and computational overhead. In this paper, we propose SimVPv2, a streamlined model that eliminates the need for Unet architectures and demonstrates that plain stacks of convolutional layers, enhanced with an efficient Gated Spatiotemporal Attention mechanism, can deliver state-of-the-art performance. SimVPv2 not only simplifies the model architecture but also improves both performance and computational efficiency. On the standard Moving MNIST benchmark, SimVPv2 achieves superior performance compared to SimVP, with fewer FLOPs, about half the training time, and 60\% faster inference efficiency. Extensive experiments across eight diverse datasets, including real-world tasks such as traffic forecasting and climate prediction, further demonstrate that SimVPv2 offers a powerful yet straightforward solution, achieving robust generalization across various spatiotemporal learning scenarios. We believe the proposed SimVPv2 can serve as a solid baseline to benefit the spatiotemporal predictive learning community.
\end{abstract}




%
\IEEEpeerreviewmaketitle

\section{Introduction}\label{sec:introduction}

%
%
%
%

 

\IEEEPARstart{A}{} wise person can foresee the future, and so should an intelligent vision model. In recent years, spatiotemporal predictive learning has gained significant attention due to its ability to infer the future by leveraging the underlying patterns embedded in spatiotemporal data, which reflect the complex and often chaotic dynamics of the real world~\cite{convlstm,babaeizadeh2017stochastic,mim,srivastava2015unsupervised,zhang2023multi,yu2023augmented,liu2023multi,li2023fast,wen2022video,yu2023magvit,wang2024omnitokenizer,gupta2023photorealistic}. Despite its vast potential, this task presents substantial challenges due to the inherent complexity and randomness in the data, such as non-linear dynamics, long-term dependencies, and high-dimensionality. To address these challenges, numerous methods have emerged, introducing a variety of novel operators, sophisticated neural architectures, and advanced training strategies aimed at improving the accuracy and efficiency of predictions. Many of these methods achieve impressive performance gains by incorporating recurrent networks~\cite{convlstm,mim,predrnn}, transformer-based models~\cite{weissenborn2019scaling,rakhimov2020latent}, and complex autoregressive or normalizing flow models~\cite{crevnet}. Additionally, various training techniques, such as adversarial learning~\cite{mathieu2015deep}, have been adopted to improve the fidelity of generated future frames. However, these advances come at a cost: increasing model complexity. As models become more intricate, they require more computational resources, are more challenging to train and scale. This raises a fundamental question: Can a simple model without recurrent units achieve comparable performance, while offering better scalability and interpretability?

\begin{figure}[t]
\centering
\vspace{1mm}
\includegraphics[width=0.48\textwidth]{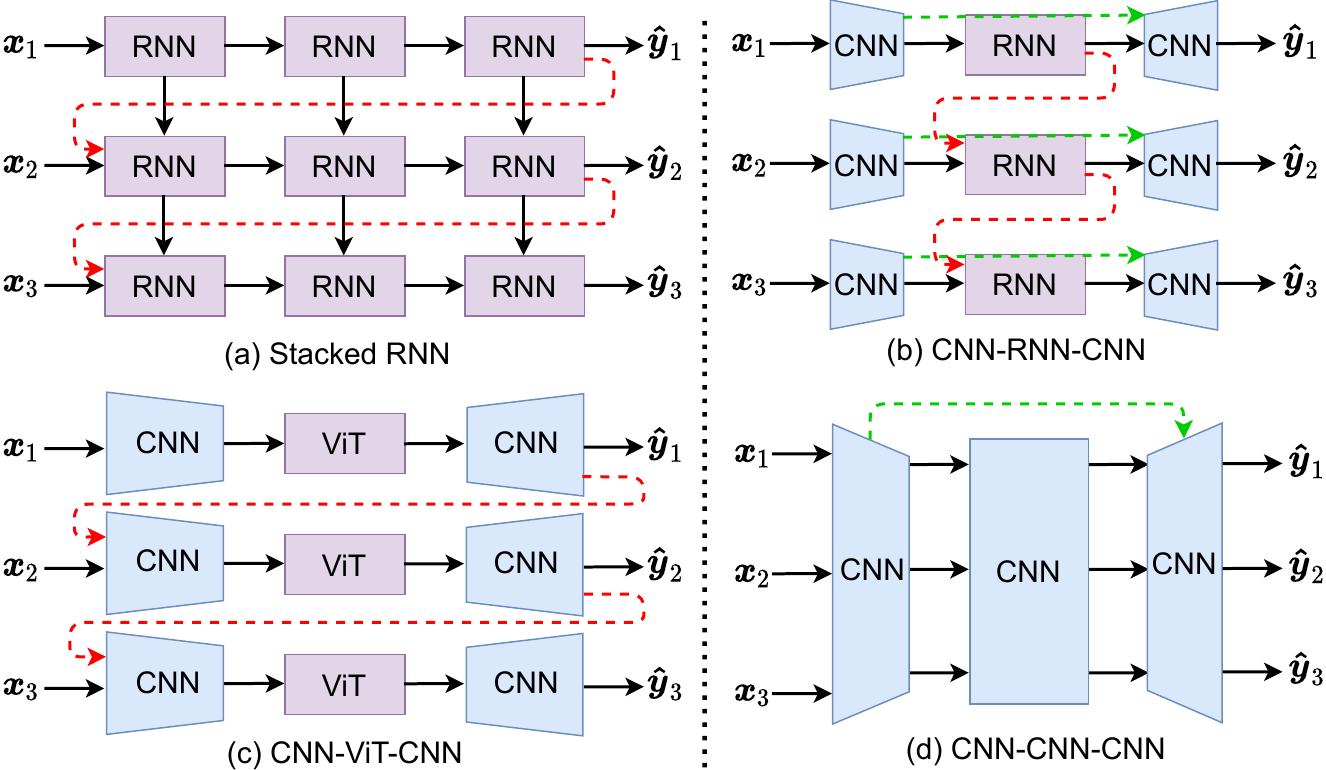}
\vspace{1mm}
\caption{Major categories of the architectures for spatiotemporal predictive learning. The red and blue dotted line are available to learn the temporal evolution and spatial dependency. Our proposed SimVP and SimVPv2 belong to (d) CNN-CNN-CNN, which can outperform other state-of-the-art methods.}
\label{fig:cmp_archi}
\end{figure}

Our previous work, SimVP, answers this question by introducing a pure convolution neural network (CNN) model that demonstrates the possibility of removing recurrent units without sacrificing predictive performance. To provide a clearer understanding of existing works in spatiotemporal predictive learning, we categorize current approaches into four broad groups, as illustrated in Fig.~\ref{fig:cmp_archi}. The first category consists of stacked recurrent neural network (RNN) models, which rely solely on recurrent units for both spatial and temporal modeling. The second group includes CNN-RNN hybrid models, which combine CNN for spatial feature extraction with RNNs for modeling temporal dynamics. The third group comprises CNN-ViT models, which integrate vision Transformer (ViT) to capture global temporal relationships using attention mechanisms, while CNNs are responsible for spatial feature extraction and reconstruction. The final category includes fully convolutional models that use CNNs for both spatial and temporal modeling and thus eliminate the need for recurrent units. SimVP belongs to this category, offering a simpler alternative by relying entirely on convolutional layers.

\begin{table*}[t]
\centering
\caption{Representative spatiotemporal predictive learning works since 2014.}
\setlength{\tabcolsep}{1.8mm}{
\begin{tabular}{ccccc}
\toprule
& (a) Stacked RNN & (b) CNN-RNN-CNN & (c) CNN-ViT-CNN & (d) CNN-CNN-CNN \\
\midrule
2014-2015 & \makecell{ConvLSTM~\cite{convlstm}, Composite LSTM\cite{srivastava2015unsupervised},\\PGP~\cite{michalski2014modeling}} & \makecell{AE-ConvLSTM-flow~\cite{patraucean2015spatio},\\ PGN~\cite{lotter2015unsupervised}} & - & GDL~\cite{mathieu2015deep}  \\
\hline
2016-2017 & \makecell{FSTN~\cite{lu2017flexible}, PredRNN~\cite{predrnn},\\Hierarchical ConvLSTM~\cite{villegas2017learning},\\RLadder~\cite{premont2017recurrent}} & \makecell{MCnet~\cite{villegas2017decomposing}, CDNA~\cite{finn2016unsupervised},\\SV2P~\cite{babaeizadeh2017stochastic}, Dual motion GAN~\cite{liang2017dual},\\DrNet~\cite{denton2017unsupervised}} &  - & \makecell{Amersfoort et al.~\cite{van2017transformation}, \\DVF~\cite{liu2017video}, EEN~\cite{henaff2017prediction}}\\
\hline
2018-2019 & \makecell{Znet~\cite{zhang2019z}, ConvLSTM-DTD~\cite{sun2019predicting},\\ dGRU~\cite{oliu2018folded}, DDPAE~\cite{hsieh2018learning},\\ PredRNN++~\cite{predrnn++}, MIM~\cite{mim}} & \makecell{
E3D-LSTM~\cite{e3dlstm}, EPVA~\cite{villegas2018hierarchical}, \\SVG-LP~\cite{denton2018stochastic}, CrevNet~\cite{crevnet},\\ hierarchical-VRNN~\cite{castrejon2019improved}} &  Weissenborn et al.~\cite{weissenborn2019scaling} & \makecell{DPG~\cite{gao2019disentangling}, PredCNN~\cite{predcnn},\\ Retrospective Cycle GAN~\cite{kwon2019predicting}} \\
\hline
2020-2024 & \makecell{PredRNNv2~\cite{predrnnv2}, MAU~\cite{chang2021mau}, \\SwinLSTM~\cite{tang2023swinlstm}} & \makecell{Hu et al.~\cite{hu2020probabilistic}, PhyDNet~\cite{phydnet}, \\FitVid~\cite{babaeizadeh2021fitvid}, STAM~\cite{chang2022stam}, \\STRPM~\cite{chang2022strpm}} & \makecell{LVT~\cite{rakhimov2020latent}, VMRNN~\cite{tang2024vmrnn}} & \makecell{G-VGG~\cite{shouno2020photo}, Chiu et al.~\cite{chiu2020segmenting}, \\ 
DMVFN~\cite{hu2023dynamic}, MMVP~\cite{zhong2023mmvp}} \\
\bottomrule 
\end{tabular}}
\vspace{-2mm}
\label{tab:previous_works}
\end{table*}

As shown in Table~\ref{tab:previous_works}, we have gathered a range of representative works in spatiotemporal predictive learning, from which we observe that recurrent-based architectures (Fig.\ref{fig:cmp_archi} a-b) have historically been the dominant choice. These architectures, particularly those based on recurrent units, have led the field for years due to their success in handling sequential data. Inspired by the success of long short-term memory (LSTM)~\cite{hochreiter1997long} in sequential modeling, ConvLSTM~\cite{convlstm} is a seminal work on the topic of spatiotemporal predictive learning that extends fully connected LSTM to convolutional LSTM. PredRNN~\cite{predrnn} proposes Spatiotemporal LSTM (ST-LSTM) units to model spatial appearances and temporal variations in a unified memory pool. This work provides insights on designing typical recurrent units for spatiotemporal predictive learning and inspires a series of subsequent works~\cite{predrnn++, byeon2018contextvp, mim, e3dlstm, predrnnv2}. PhyDNet~\cite{phydnet} combined ConvLSTM with a two-branch architecture involving PhyCells, which incorporated physical dynamics through partial differential equations, helping guide the model with physics-based constraints. Similarly, CrevNet~\cite{crevnet} proposed an invertible two-way autoencoder based on normalizing flow, introducing a conditionally reversible architecture that allowed better information preservation across time steps. Despite the success of these recurrent-based architectures, they face several inherent limitations. The sequential nature of RNNs poses significant computational challenges, particularly for long-term predictions. As each time step is processed sequentially, computations cannot be fully parallelized, leading to inefficiencies in training and inference. These limitations motivate the exploration of simpler and more efficient architectures that can handle spatiotemporal dynamics without the computational overhead of recurrence.

In contrast, purely CNN-based models (Fig.~\ref{fig:cmp_archi} d) are not as favored as the above RNN-based approaches (Fig.~\ref{fig:cmp_archi} a-b). Moreover, the existing methods usually require fancy techniques, e.g., adversarial training~\cite{kwon2019predicting}, teacher-student distilling~\cite{chiu2020segmenting}, and optical flow~\cite{gao2019disentangling}. In an effort to simplify the landscape of spatiotemporal predictive learning, SimVP was introduced as a fully convolutional architecture that used common components such as convolutional networks for both spatial and temporal modeling, simple shortcut connections for efficient feature propagation and was trained end-to-end with mean squared error loss. SimVP proved to be successful in demonstrating that pure CNN with Unet-insipred multi-scale processing could achieve comparable performance.

Despite the success of SimVP, however, the reliance on Unet architectures introduced its own set of challenges. These multi-scale processing frameworks, with their top-down and bottom-up paths and skip connections, are still computationally expensive and complex. While SimVP streamlined many aspects of spatiotemporal predictive learning, the complexity of the Unet structure limited its efficiency, particularly in terms of computation and model simplicity. 

Building on the foundation laid by SimVP, we propose SimVPv2, a further simplified model that completely eliminates the need for Unet architectures. Instead of relying on complex multi-scale and skip-connection frameworks, SimVPv2 introduces a novel and efficient Gated Spatiotemporal Attention (gSTA) mechanism. The gSTA module captures both spatial and temporal dependencies without the overhead of hierarchical or multi-branch processing, achieving a streamlined architecture that is more computationally efficient. As a result, SimVPv2 significantly reduces the number of parameters and FLOPs, leading to shorter training times and increased inference efficiency, while maintaining or improving predictive accuracy. Extensive experiments across diverse datasets demonstrate that SimVPv2 not only surpasses SimVP in terms of performance but also generalizes effectively across a wide range of spatiotemporal predictive tasks.

A preliminary version of this work was published in~\cite{simvp}. This journal paper extends it in the following aspects: 
\begin{itemize}
\item We introduce SimVPv2, a more streamlined and efficient architecture that completely removes the need for Unet structures, replacing them with gSTA modules to capture both spatial and temporal dependencies with greater computational efficiency.
\item We reproduce the mainstream spatiotemporal predictive learning methods into a unified framework and systematically evaluate performance on the standard benchmark Moving MNIST dataset in consideration of computational cost and time complexity.
\item We conduct a comprehensive evaluation across a broader set of eight benchmark datasets, including both synthetic and real-world tasks. This comprehensive assessment demonstrates the robustness and generalization capabilities of the proposed approach across various spatiotemporal predictive learning challenges.
\end{itemize}
We release our code at \href{https://github.com/chengtan9907/OpenSTL/tree/SimVPv2}{github.com/chengtan9907/SimVPv2}.

\section{Related work}

\subsection{Stacked RNN} As shown in Fig.~\ref{fig:cmp_archi} (a), stacked RNN architectures have been widely adopted for spatiotemporal predictive tasks. These methods typically involve the design of novel recurrent units (local) and sophisticated architectural frameworks (global) to model temporal dependencies in sequential data. Recurrent Grammar Cells~\cite{michalski2014modeling} stacks multiple gated autoencoders in a recurrent pyramid structure. ConvLSTM~\cite{convlstm} extends fully connected LSTMs to have convolutional computing structures to capture spatiotemporal correlations. PredRNN~\cite{predrnn} suggests simultaneously extracting and memorizing spatial and temporal representations. PredRNN++~\cite{predrnn++} proposes a gradient highway unit to alleviate the gradient propagation difficulties for capturing long-term dependency. MIM~\cite{mim} uses a self-renewed memory module to model both the non-stationary and stationary properties of the video. dGRU~\cite{oliu2018folded} shares state cells between encoder and decoder to reduce the computational and memory costs. Due to the excellent flexibility and accuracy, these methods play fundamental roles in spatiotemporal predictive learning. PredRNNv2~\cite{predrnnv2} extends PredRNN by introducing a decoupling loss and a reverse scheduled sampling method. MAU~\cite{chang2021mau} leverages an attention mechanism to capture the correlations between the current spatial state and historical states, aggregating motion and appearance to improve predictions. SwinLSTM~\cite{tang2023swinlstm} merges the strengths of Swin Transformer blocks with a simplified LSTM architecture and replaces the convolutional structure of ConvLSTM with the self-attention mechanism from transformers.

\subsection{CNN-RNN-CNN} This framework projects video frames to the latent space and employs RNN to predict the future latent states, seeing Fig.~\ref{fig:cmp_archi} (b). In general, they focus on modifying the LSTM and encoding-decoding modules. Spatio-Temporal video autoencoder~\cite{patraucean2015spatio} incorporates ConvLSTM and an optical flow predictor to capture changes over time. Conditional VRNN~\cite{castrejon2019improved} combines CNN encoder and RNN decoder in a variational generating framework. E3D-LSTM~\cite{e3dlstm} applies 3D convolution for encoding and decoding and integrates it into latent RNNs for obtaining motion-aware and short-term features. CrevNet~\cite{crevnet} proposes using CNN-based normalizing flow modules to encode and decode inputs for information-preserving feature transformations. PhyDNet~\cite{phydnet} models physical dynamics with CNN-based PhyCells. Recently, this framework has attracted considerable attention because the CNN encoder can extract compressed features for accurate and efficient prediction. FitVid~\cite{babaeizadeh2021fitvid} employs a conditional variational model, using LSTMs to predict hidden states in a probabilistic manner. STAM~\cite{chang2022stam} utilizes 3D convolutional layers in recurrent units to jointly learn high-level semantic features and low-level texture representations. STRPM~\cite{chang2022strpm} introduces Residual Predictive Memory (RPM), which focuses on modeling the spatiotemporal residuals between consecutive frames. Additionally, STRPM is trained with generative adversarial networks and incorporates a learned perceptual loss, improving the perceptual quality of the generated frames.

\subsection{CNN-ViT-CNN} This framework introduces Vision Transformer (ViT) to model latent dynamics. By extending language transformer~\cite{vaswani2017attention} to ViT~\cite{dosovitskiy2020image}, a wave of research has been sparked recently. DeiT~\cite{touvron2021training} and Swin Transformer~\cite{liu2021swin} have achieved superior performance on various computer vision tasks. The great success of image transformers has inspired the investigation of video transformers. VTN~\cite{neimark2021video} applies sliding window attention on temporal dimension following a 2D spatial feature extractor. TimeSformer~\cite{bertasius2021space} and ViViT~\cite{arnab2021vivit} explore various strategies for space-time attention. MViT~\cite{fan2021multiscale} introduces a multiscale pyramid feature extractor, capturing both fine-grained and high-level temporal patterns across varying resolutions. Video Swin Transformer~\cite{liu2021video} extend Swin Transformer from 2D to 3D, utilizing shiftable local attention windows to balance speed and accuracy. VMRNN~\cite{tang2024vmrnn} proposes the VMRNN cell, a recurrent unit that integrates the strengths of Vision Mamba~\cite{zhuvision} blocks with LSTM. Many of these works focus on video generation using ViTs~\cite{weissenborn2019scaling,rakhimov2020latent}. For instance, MAGViT~\cite{yu2023magvit} introduces a 3D tokenizer that quantizes video into spatiotemporal tokens and utilizes non-autoregressive decoding to generate tokens. OmniTokenizer~\cite{wang2024omnitokenizer} trains on fixed-resolution image data to develop strong spatial encoding capabilities and then jointly trains on both image and video data at multiple resolutions to learn temporal dynamics.

\vspace{-1mm}
\subsection{CNN-CNN-CNN} This framework is not as popular as the previous one because it is so simple that complex modules and training strategies are required to improve performance. DVF~\cite{liu2017video} learns the voxel flow by an autoencoder to reconstruct a frame by borrowing voxels from nearby frames. DMVFN~\cite{hu2023dynamic} proposes a routing module to dynamically select a sub-network. PredCNN~\cite{predcnn} combines cascade multiplicative units with CNN to capture inter-frame dependencies. DPG~\cite{gao2019disentangling} disentangles motion and background via a flow predictor and a context generator. G-VGG~\cite{shouno2020photo} uses a hierarchical model to make predictions at different scales and train the model with adversarial and perceptual loss. Chiu et al.~\cite{chiu2020segmenting} encodes RGB frames from the past and decodes the future semantic segmentation using teacher-student distilling. MMVP~\cite{zhong2023mmvp} decouples motion and appearance information by constructing appearance-agnostic motion matrices.

\vspace{-1mm}
\subsection{Summary}

Although mainstream reucrrent-based approaches have made significant progress, the reliance on recurrent units still imposes considerable computational overhead, and their architectures face challenges in parallelization. ViT-based models, while powerful, demand substantial computational resources. Fully CNN-based models offer a more efficient solution but depend on complex techniques. SimVP represents an important step in this direction, demonstrating the potential of fully convolutional architectures. However, its reliance on Unet-like structures remains a source of computational complexity. SimVPv2 builds on the simplicity of fully CNN-based models, taking it a step further by removing the need for complex components such as Unet, recurrent units, or transformers.

\begin{figure*}[ht]
\centering
\includegraphics[width=1.0\textwidth]{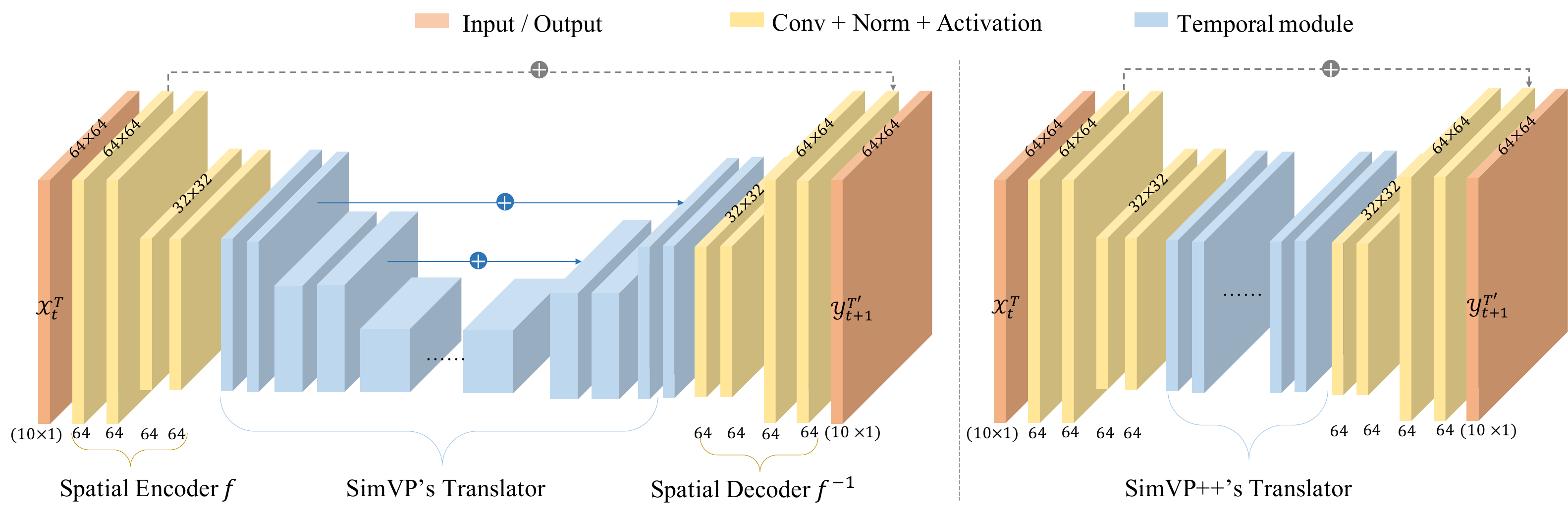} 
\caption{The overall framework of SimVP and SimVPv2.}
\vspace{-2mm}
\label{fig:overview} 
\end{figure*}  

\section{Preliminaries}

We formally define the spatiotemporal predictive learning problem as follows. Given a video sequence $\mathcal{X}^{t, T} = \{\mathbf{x}^i\}_{t-T+1}^t$ at time $t$ with the past $T$ frames, we aim to predict the subsequent $T'$ frames $\mathcal{Y}^{t+1, T'} = \{\mathbf{x}^{i}\}_{t+1}^{t+T'}$ from time $t+1$, where $\mathbf{x}_i \in \mathbb{R}^{C \times H \times W}$ is usually an image with channels $C$, height $H$, and width $W$. In practice, we represent the input observed sequences and output predicted sequences as tensors, i.e.,  $\mathcal{X}^{t, T} \in \mathbb{R}^{T \times C \times H \times W}$ and $\mathcal{Y}^{t+1, T'} \in \mathbb{R}^{T' \times C \times H \times W}$.

The model with learnable parameters $\Theta$ learns a mapping $\mathcal{F}_\Theta: \mathcal{X}^{t, T} \mapsto \mathcal{Y}^{t+1, T'}$ by exploring both spatial and temporal dependencies. In our case, the mapping $\mathcal{F}_\Theta$ is a neural network model trained to minimize the difference between the predicted future frames and the ground-truth future frames. The optimal parameters $\Theta^*$ are:
\begin{equation}
  \Theta^* = \arg\min_{\Theta} \mathcal{L}(\mathcal{F}_\Theta(\mathcal{X}^{t, T}), \mathcal{Y}^{t+1, T'}),
\end{equation}
where $\mathcal{L}$ is a loss function that evaluates such differences.

\section{Method}

\subsection{Motivation}

\begin{figure}[htbp]
\centering
\includegraphics[width=0.49\textwidth]{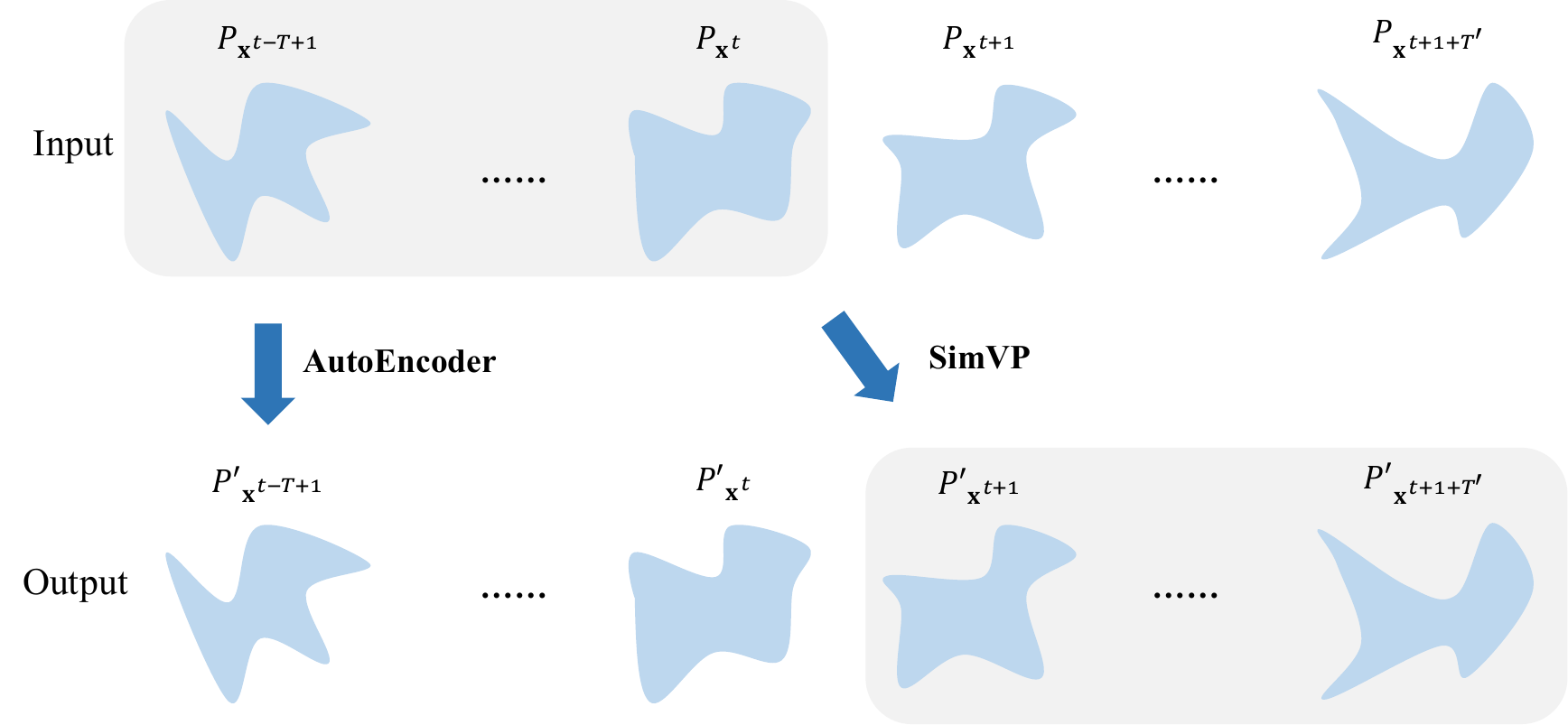} 
\caption{The schematic diagram of the autoencoder and our proposed SimVP. While the autoencoder focuses on a single frame at a static time, SimVP concerns a sequence of frames at a dynamic time. The first row denotes the ground-truth frames, and the second denotes the predicted frames. From left to right, the data changes over time.}
\label{fig:cmp_ae}
\end{figure}

Inspired by the autoencoder that reconstructs a single frame image and captures spatial correlations, we aim to design an autoencoder-like architecture that inputs the past frames and outputs the future frames while preserving the temporal dependencies. As shown in Fig.~\ref{fig:cmp_ae}, the traditional autoencoder focuses on single frame image reconstruction at a static time and learns a mapping $\mathcal{G}_\Phi: \mathbf{x} \mapsto \mathbf{x}$ to minimize the divergence between the decoded output probability distribution $P'_{\mathbf{x}} = \mathcal{G}_\Phi(\mathbf{x})$ and the encoded input probability distribution $P_{\mathbf{x}}$. Its optimal parameters $\Phi^*$ are:
\begin{equation}
\Phi^* =\arg\min_\Phi Div(P_{\mathbf{x}}, P'_{\mathbf{x}}),
\end{equation}
where $Div$ denotes a specific divergence measure. In practice, we usually minimize the MSE loss between $\mathbf{x}$ and $\mathcal{G}_\Phi(\mathbf{x})$ as follows:
\begin{equation}
  \Phi^* =\arg\min_\Phi \|\mathbf{x} - \mathcal{G}_\Phi(\mathbf{x})\|^2.
\end{equation}

Similar to the autoencoder, SimVP learns a mapping $\mathcal{F}_\Theta: \mathcal{X}^{t, T} \mapsto \mathcal{Y}^{t+1, T'}$ to encode the past frames $\mathcal{X}^{t, T}$ and decode the future frames $\mathcal{Y}^{t+1, T'}$ and thus extends the autoencoder-like framework along the time axis. The optimal parameters $\Theta^*$ are:
\begin{equation}
\Theta^* =\arg\min_\Theta \sum_{t+1}^{t+1+T'} Div(P_{\mathbf{x}^i}, P'_{\mathbf{x}^i}).
\end{equation}
Analogous to the autoencoder, we minimize the MSE loss between $\mathbf{x}^i$ and $\mathcal{F}_\Theta(\mathbf{x}^i)$ in practice:
\begin{equation}
  \Theta^* =\arg\min_\Theta \sum_{t+1}^{t+1+T'} \|\mathbf{x}^i - \mathcal{F}_\Theta(\mathbf{x}^i)\|^2.
\end{equation}

\subsection{Overview}

Taking input Moving MNIST data as an example, we provide an overview of SimVP and SimVPv2 models, as illustrated in Fig.~\ref{fig:overview}. They share a similar architecture but SimVPv2 introduces a significantly streamlined design which eliminates the Unet multi-scale processing and complex hierarchical structures. The spatial encoder is employed to encode the high-dimensional past frames into the low-dimensional latent space, and the translator learns both spatial dependencies and temporal variations from the latent space. The spatial decoder decodes the latent space into the predicted frames. 

Striving for simplicity, We implement the spatial encoder with $N_s$ vanilla convolutional layers ('Conv2d' in PyTorch) and the spatial decoder with $N_s$ upsampling layers ('ConvTranspose2d' or 'PixelShuffle' in PyTorch). The hidden representations in the spatial encoder $f$ are as follows:
\begin{equation}
\begin{aligned}
z_i = \sigma(\textrm{Norm2d}(\textrm{Conv2d}(z_{i-1}))), 1 \leq i \leq N_s,
\end{aligned}
\end{equation}
where $\sigma$ is a nonlinear activation, $\textrm{Norm2d}$ is a normalization layer, $z_0$ is the input tensor. The strides of the convolutional layers are one, except downsampling, which has a stride of two. For every two convolutional layers, we perform downsampling once. The hidden representations in the spatial decoder $f^{-1}$ can be formally described as:
\begin{equation}
\begin{aligned}
z_k = \sigma(\textrm{Norm2d}(\textrm{unConv2d}(z_{i-1}))), \\ N_s+N_t < k \leq 2N_s+N_t,
\end{aligned}
\end{equation}
where $\textrm{unConv2d}$ is a transposed convolutional layer or pixelshuffle layer if it needs upsampling. Otherwise, it is a convolutional layer with stride one.

The middle spatiotemporal translator of the model consists of $N_t$ temporal modules, which we illustrate in detail in Section~\ref{sec:translator}. The hidden representations in this part are:
\begin{equation}
z_j = \textrm{TemporalModule}(z_{i-1}), \\ N_s < j \leq N_s+N_t,
\end{equation}
where $z_{N_s-1}$ is the output of the spatial encoder. A residual connection from the first layer in the spatial encoder to the last layer in the spatial decoder is introduced to preserve the spatial feature. The mapping $\mathcal{F}_\Theta$ is the composition of the above components:
\begin{equation}
  \mathcal{F}_\Theta = f^{-1} \circ h \circ f
\end{equation}

\begin{figure}[htbp]
\centering
\includegraphics[width=0.48\textwidth]{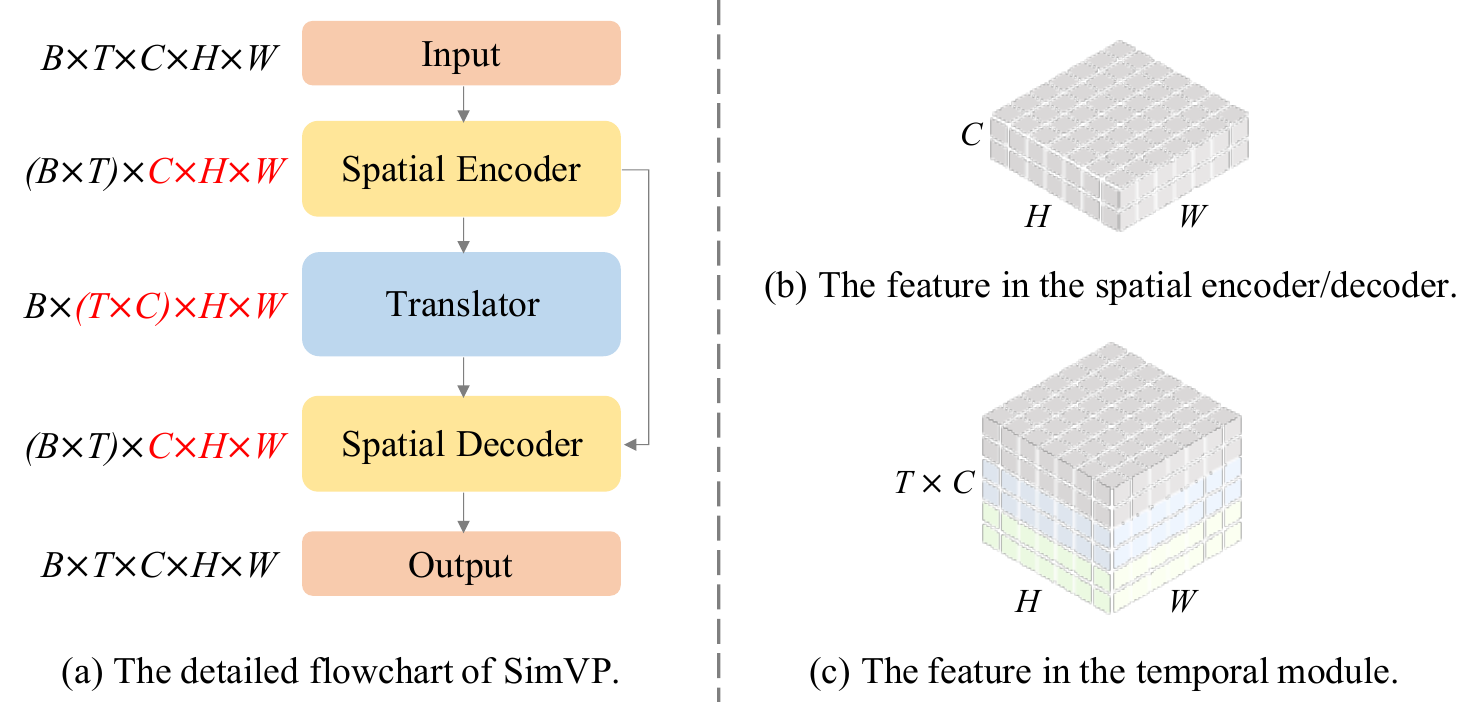} 
\caption{The spatial encoder and decoder perform single-frame level spatial feature extraction and reconstruction. The translator learns from multi-frame level temporal dependencies.}
\label{fig:flow_chart} 
\end{figure}

\begin{figure*}[htbp]
\centering
\includegraphics[width=0.99\textwidth]{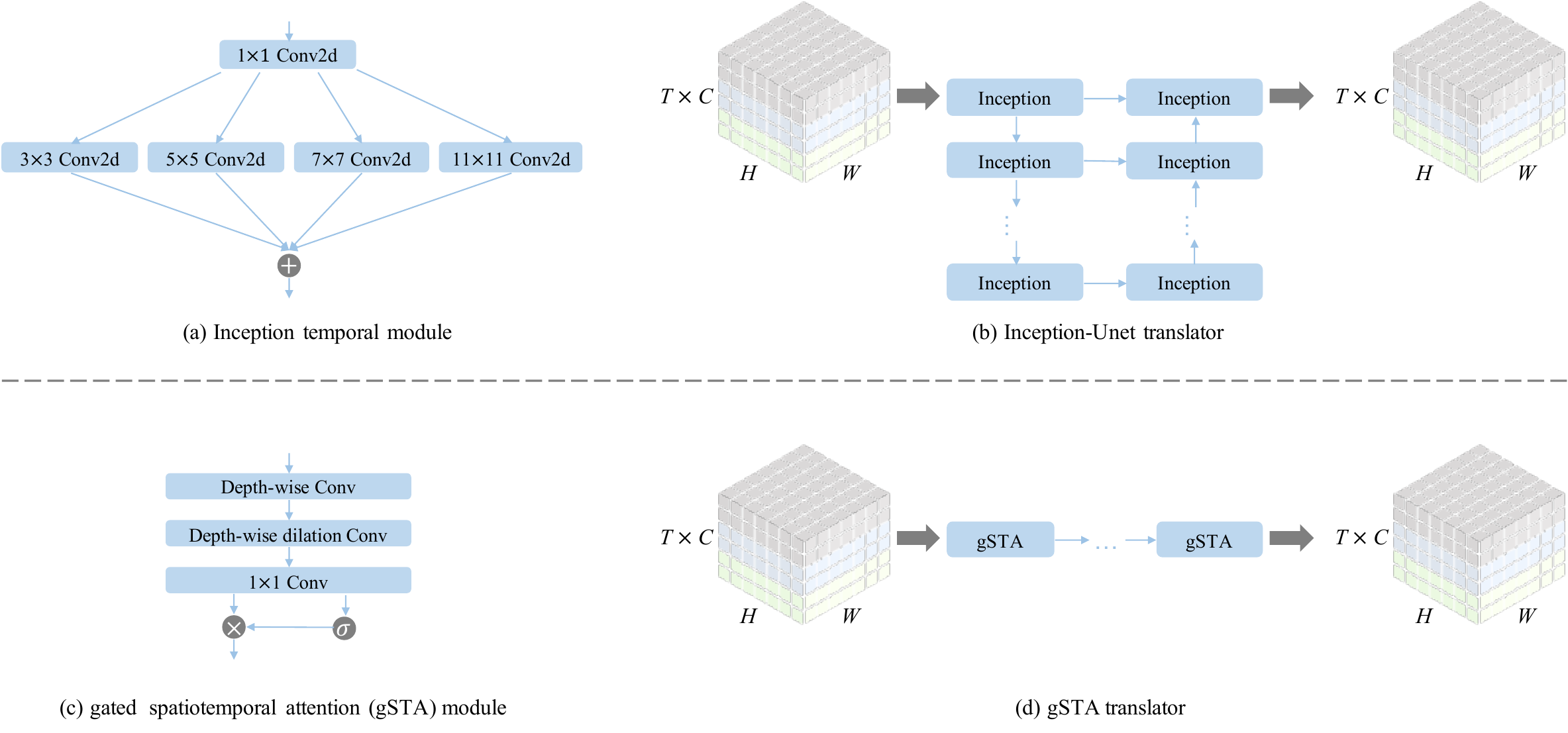} 
\caption{(a-b) The Inception temporal module and corresponding Inception-Unet translator. (c-d) The gSTA module and corresponding gSTA translator.}
\label{fig:module} 
\end{figure*}

Given a batch of input past frames $\mathcal{B} \in \mathbb{R}^{B\times T\times C\times H\times W}$ with the batch size of $B$. In the spatial encoder and decoder, we reshape the input tensors into tensors of shape $(B\times T)\times C\times H\times W$, as shown in Fig.~\ref{fig:flow_chart} (b). Thus, the spatial encoder and decoder treat each frame as a single sample and focus on the single-frame level features regardless of the temporal variations. In the translator, we reshape the hidden representations from the spatial encoder into tensors of shape $B\times (T\times C)\times H\times W$ and stack multi-frame level features along the time axis, as shown in Fig.~\ref{fig:flow_chart} (c). By forcing the designed temporal module built upon convolutional networks to learn from stacks of multi-frame features, SimVP can capture the intrinsic temporal evolutions inside the sequential data.
  
\subsection{Spatiotemporal Translator}
\label{sec:translator}

The spatiotemporal translator takes the encoded hidden representations of the spatial encoder $f$ as input and outputs hidden spatiotemporal representations for the spatial decoder $f^{-1}$ to decode. Here, we introduce two kinds of spatiotemporal translators built upon pure convolutional neural networks.

\subsubsection{Inception-Unet Translator}

In the conference version of SimVP, we design an Inception-like temporal module and build the middle spatiotemporal translator with blocks of this module. 

As shown in Fig.~\ref{fig:module} (a), our Inception temporal module is different from the original Inception module~\cite{szegedy2015going} in the following aspects: (1) We apply $1\times 1$ convolution at the front instead of at the end for increasing the hidden dimension in advance. This operation is not responsible for better performance but convenience. (2) We employ larger kernels (e.g., $7\times 7$ and $11 \times 11$) than the vanilla Inception module. Larger kernels are preferred for globally distributed information, while smaller kernels are preferred for locally distributed information. Spatiotemporal predictive learning usually faces the difficulty of considerable variations in the location of the valuable information along with time. By leveraging such a multi-branch architecture, the Inception temporal module can jointly obtain both local and global features from stacks of temporal dynamics. (3) The output features from convolutional layers with different kernel sizes are added up instead of concatenated as the simplicity of keeping the same dimension. Our Inception temporal module can be formally described as:
\begin{equation}
  \hat{z}^{j} = \mathrm{Conv2d}_{1\times 1}(z^{j}),
\end{equation}
\begin{equation}
  {z}^{j+1} = \sum_{k \in \{3, 5, 7, 11\}}\mathrm{Conv2d}_{k\times k}(\hat{z}^{j}),
\end{equation}
The middle spatiotemporal translator is built based on the above Inception modules with an Unet-like architecture. The input hidden representations are firstly passed through several Inception temporal modules from top to bottom and then go through a symmetric path from bottom to top. We have concatenation connections between the top-down and bottom-up paths for every Inception temporal module. Note that there is no contracting in the top-down path and expanding in the bottom-up path for simplicity, which is different from the vanilla Unet~\cite{ronneberger2015u}.

\subsubsection{Gated Spatiotemporal Attention Translator}
In this journal version, we propose a gated spatiotemporal attention module and build the middle spatiotemporal translator by stacking such modules instead of using Unet architecture, which further simplifies the model in both time and space complexity.  Though this module is still built on pure convolutional networks, it is efficient in capturing spatiotemporal dependencies.

Attention mechanism, which is a hotspot in visual transformers, can adaptively select discriminative features and ignore noisy responses according to the input features. We aim to design a spatiotemporal attention module that automatically captures features relying on temporal dependencies and spatial correlations. Recent research has revealed that large kernel convolutions share advantages with vision transformers in obtaining large effective receptive fields and higher shape bias rather than texture bias~\cite{ding2022scaling,liu2022convnet,liu2021pay,guo2022visual}. Motivated by this observation, we leverage large kernel convolutions to imitate the attention mechanism and extract spatiotemporal attention from the input representations in the latent space. 

\begin{figure}[ht]
  \centering
  \includegraphics[width=0.49\textwidth]{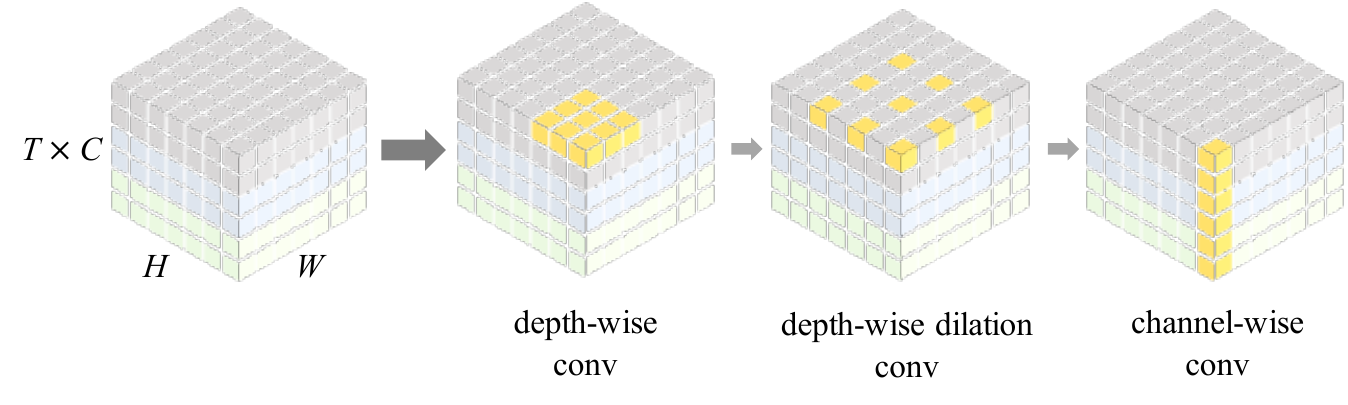} 
  \caption{The large kernel convolution in the gated spatiotemporal attention module. The yellow region denotes the receptive field.}
  \label{fig:large_kernel} 
  \end{figure}

However, directly utilizing large kernel convolutions suffers from inefficient computation and a huge amount of parameters. As an alternative, we decompose the large kernel convolution~\cite{liu2022convnet,ding2022scaling,guo2022visual} into several components: (1) a depth-wise convolution that captures local receptive fields within a single channel, (2) a depth-wise dilation convolution that builds connections between distant receptive fields, (3) a $1 \times 1$ convolution that performs channel-wise interactions. A $(2d - 1) \times (2d - 1)$ depth-wise convolution and a $\frac{K}{d} \times \frac{K}{d}$ depth-wise dilation convolution with dilation $d$ have a receptive field with a size of $K \times K$, and a channel-wise $1 \times 1$ further assist them in multi-channel connections. We use the above three components to simulate the large kernel convolution with a low computational overhead, as shown in Fig.~\ref{fig:large_kernel}.

The gated spatiotemporal attention (gSTA) module is illustrated in Fig.~\ref{fig:module} (c). Benefited by the large receptive fields, we can capture long-range correlations in both spatial and temporal perspectives. We split the output of the above large kernel convolution operation into two parts and take one of them with a sigmoid function as an attention gate. The gated spatiotemporal attention module is formalized as:
\begin{equation}
  \hat{z}^{j} = \mathrm{Conv2d}_{1\times 1}(\mathrm{Conv}_{\mathrm{Dw-d}}(\mathrm{Conv}_{\mathrm{Dw}}(z^{j}))),
\end{equation}
\begin{equation}
  g, \bar{z}^{j} = \mathrm{split}(\hat{z}^{j}),
\end{equation}
\begin{equation}
  z^{j+1} = \sigma(g) \odot \bar{z}^{j},
\end{equation}
where $\mathrm{Conv}_{\mathrm{Dw}}$ is the depth-wise convolution and $\mathrm{Conv}_{\mathrm{Dw-d}}$ is the depth-wise dilation convolution, $g$ is the attention coefficients, and $\odot$ denotes element-wise multiplication.

We show the gSTA translator in Fig.~\ref{fig:module} (d). With the gSTA module in place, we can build the middle translator by simply stacking several gSTA modules without Unet architecture. The spatiotemporal attention coefficient $g$ provides a dynamic mechanism that adaptively changes according to the input features. The gated attention $\sigma(g)$ gate is used to adaptively select the informative features and filter unimportant features from a spatiotemporal perspective.

\subsection{Advantages of gSTA over IncepU}

One key advantage of gSTA is the \textbf{enhanced receptive field coverage} achieved through the use of large kernel convolutions and dilated convolutions, enabling the module to capture long-range dependencies more effectively. Another important benefit of gSTA is its ability to \textbf{dynamically select informative features through the gating mechanism}, which acts as an adaptive filter that enhances significant features while suppressing less relevant ones, allowing the model to focus on the most salient spatiotemporal patterns.
Moreover, gSTA offers \textbf{reduced computational complexity and improved parameter efficiency}. Unlike IncepU's multi-branch architecture, which introduces multiple convolutional pathways and increases the computational burden, gSTA simplifies the architecture by employing a single unified path with large kernel convolutions and gating mechanisms.

\begin{table*}[ht]
\centering
\caption{Quantitative results of different methods on the Moving MNIST dataset ($10 \rightarrow 10$ frames). Note that "-S" denotes the smaller model and "-L" denotes the larger model. We report "SimVPv2-S$\times 3$" by training "SimVPv2-s" with three times as much epoch, i.e., 600 epochs. Its training efficiency is reported by multiplying the original time by three.}
\vspace{-3mm}
\setlength{\tabcolsep}{3.9mm}{
\begin{tabular}{lcccccc}
\toprule
Method & FLOPs (G) $\downarrow$ & Training time $\approx$ (s) $\downarrow$ & Inference efficiency $\uparrow$ & MSE $\downarrow$ & MAE $\downarrow$ & SSIM $\uparrow$ \\
\midrule
ConvLSTM-S & \textbf{14.45} & 190 & 7.50 & 46.26$\pm$0.26 & 142.18$\pm$0.61 & 0.878$\pm$0.001 \\
PhyDNet    & \underline{15.33} & \underline{452} & 4.62 & 35.68$\pm$0.40 & 96.70$\pm$0.29  & 0.917$\pm$0.000 \\
MAU        & 17.79 & 535 & 3.08 & 30.64$\pm$0.10 & 88.17$\pm$0.35  & 0.928$\pm$0.001 \\
\rowcolor{gray!30} SimVP & 19.43 & \underline{261} & \underline{27.15} & 32.22$\pm$0.02 & 89.19$\pm$0.33 & 0.927$\pm$0.000  \\
\rowcolor{gray!30} SimVPv2-S & \underline{16.53} & \textbf{156} & \textbf{44.09} & \textbf{26.60}$\pm$0.02 & \textbf{77.32}$\pm$0.22 & \textbf{0.940}$\pm$0.000 \\
\hline
\hline
ConvLSTM-L & 127.01 & 879  & 6.24 & 29.88$\pm$0.17 & 95.05$\pm$0.25 & 0.925$\pm$0.000 \\
PredRNN    & 115.95 & 869  & 3.97 & 25.04$\pm$0.08 & 76.26$\pm$0.29 & 0.944$\pm$0.000 \\
PredRNN++  & 171.73 & 1280 & 3.71 & 22.45$\pm$0.36 & 69.70$\pm$0.25 & 0.950$\pm$0.000 \\
MIM        & 179.18 & 1388 & 3.08 & 23.66$\pm$0.20 & 74.37$\pm$0.46 & 0.946$\pm$0.000 \\
E3D-LSTM & 298.87 & 2693 & 3.73 & 36.19$\pm$0.20 & 78.64$\pm$0.35 & 0.932$\pm$0.000 \\
CrevNet & 270.68 & 1166 & 1.01 & 30.15$\pm$1.61 & 86.28$\pm$2.65 & 0.935$\pm$0.003\\
PredRNNv2 & 116.59 & 899 & 3.49 & 27.73$\pm$0.08 & 82.17$\pm$0.33 & 0.937$\pm$0.000 \\
SwinLSTM  & 69.87 & 820 & 6.51 & 27.44$\pm$0.08 & 78.69$\pm$0.29 & 0.938$\pm$0.000 \\
MMVP      & 93.55 & 402 & 21.33 & 33.29$\pm$0.02 & 89.61$\pm$0.23 & 0.926$\pm$0.000 \\
\rowcolor{gray!30} SimVPv2-S$\times 10$ & \underline{16.53} & 1560 & \textbf{44.09} & \textbf{15.05}$\pm$0.03 & \textbf{49.80}$\pm$0.10 &\textbf{0.967}$\pm$0.000 \\
\rowcolor{gray!30} SimVPv2-S$\times 5$ & \underline{16.53} & 780 & \textbf{44.09} & \underline{16.47}$\pm$0.02 & \underline{53.24}$\pm$0.04 & \underline{0.964}$\pm$0.000\\
\rowcolor{gray!30} SimVPv2-S$\times 3$ & \underline{16.53} & \underline{468} & \textbf{44.09} & 22.37$\pm$0.06 & 67.52$\pm$0.03 & 0.951$\pm$0.000\\
\rowcolor{gray!30} SimVPv2-L & 152.20 & 796 & \underline{21.23} & 21.81$\pm$0.03 & 66.43$\pm$0.04 & 0.952$\pm$0.000 \\
\bottomrule
\end{tabular}
}
\vspace{-4mm}
\label{tab:mmnist}
\end{table*}

\section{Experiments}

The experiments are conducted on various datasets with different settings to evaluate from the following aspects:
\begin{itemize}
\item Standard spatiotemporal predictive learning (Section~\ref{lab:standard}). We regard the video prediction problem with the same number of input and output frames as the standard spatiotemporal predictive learning. We evaluate the performance on \textbf{Moving MNIST~\cite{srivastava2015unsupervised}, TaxiBJ~\cite{zhang2017deep}, and WeatherBench~\cite{rasp2020weatherbench}}.
\item Generalization ability across different datasets (Section~\ref{lab:generalization}). Generalizing the learned knowledge to other domains is a challenge. We investigate such ability by training the model on the \textbf{KITTI~\cite{geiger2013vision}} and evaluating it on the \textbf{Caltech Pedestrian~\cite{dollar2009pedestrian}.}
\item Predicting frames with flexible lengths (Section~\ref{lab:flexible}). One of the advantages of recurrent units is that they can easily handle flexible-length frames like the \textbf{KTH}~\cite{schuldt2004recognizing}. Our work tackles the long-length frame prediction by imitating recurrent units that feed predicted frames as the input and recursively produce long-term predictions.
\item Challenging multi-domain evaluation (Section~\ref{lab:challenging}). \textbf{RoboNet}~\cite{dasari2019robonet} is designed for robotic action planning, containing a diverse collection of robot interactions across various environments. \textbf{BridgeData}~\cite{walke2023bridgedata,wang2024predbench} is a multi-domain dataset including 71 distinct tasks across 10 different scenes. They pose significant challenges due to their need for adapting to various environments.

\end{itemize}

\begin{figure*}[ht]
\centering
\includegraphics[width=0.98\textwidth]{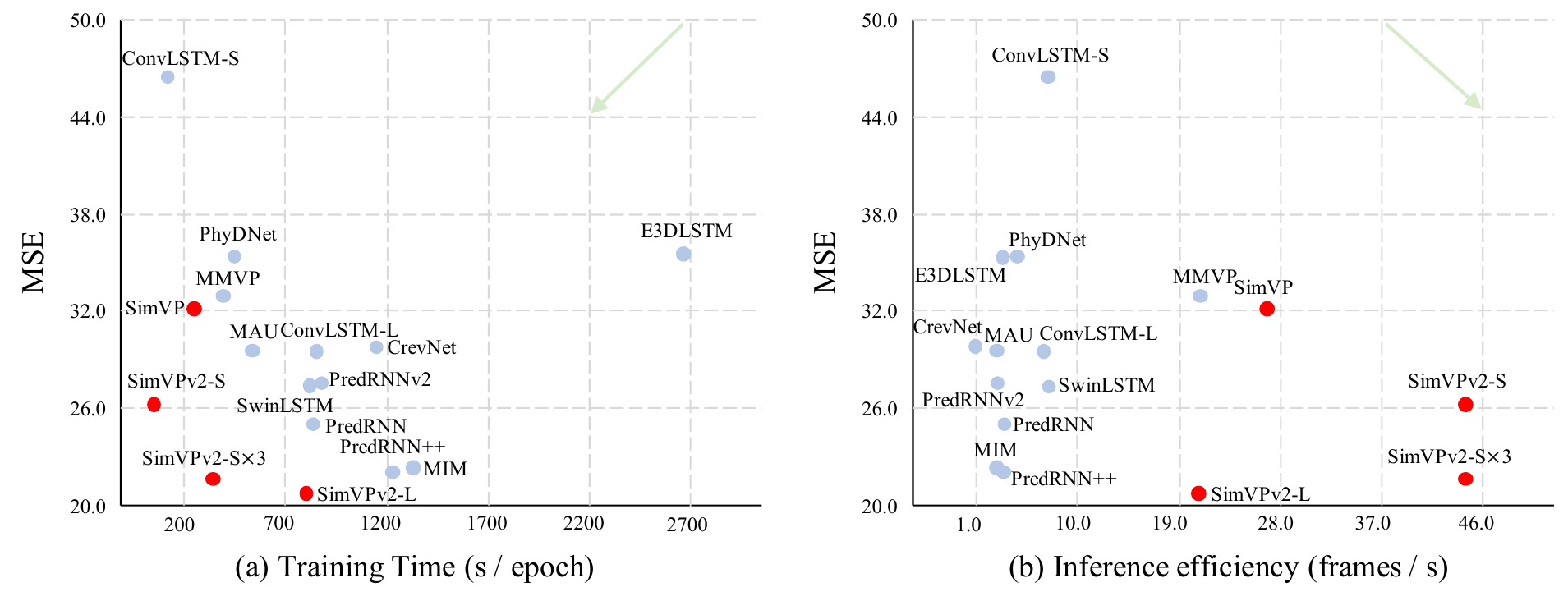} 
\vspace{-2mm}
\caption{The performance of SimVPs on the Moving MNIST dataset. The variants of SimVP are denoted in red color. For the training time, the less the better. For the inference efficiency (frames per second), the more the better. The light green arrow indicates the direction of model optimization.}
\vspace{-2mm}
\label{fig:train_inference_efficiency} 
\end{figure*}

\vspace{-2mm}
\subsection{Standard spatiotemporal predictive learning}

\label{lab:standard}
\subsubsection{Moving MNIST}
In this dataset, each video is generated 20 frames long and consists of two digits inside a $64 \times 64$ patch. The digits are randomly selected and placed initially at random locations. Each digit is assigned a velocity whose direction is chosen uniformly at random on the unit circle and whose size is chosen uniformly at random within a fixed range. The digits bounce off the edges of the 64$\times$64 frame and overlap if they are in the same position.

We evaluate state-of-the-art methods with the same protocol for fair comparisons, including  ConvLSTM~\cite{convlstm}, PredRNN~\cite{predrnn}, PredRNN++~\cite{predrnn++}, MIM~\cite{mim}, E3D-LSTM~\cite{e3dlstm}, PhyDNet~\cite{phydnet}, CrevNet~\cite{crevnet}, MAU~\cite{chang2021mau}, SwinLSTM~\cite{tang2023swinlstm}, and MMVP~\cite{zhong2023mmvp}. By using ConvLSTM with different sizes as the standard baselines, we divide these models into two groups according to their computational cost, as shown in Table~\ref{tab:mmnist}. ConvLSTM-S is the small model with four layers with a hidden size of $64$, and ConvLSTM-L is the large model with four layers with a hidden size of $192$. Models are trained using the Adam optimizer~\cite{kingma2015adam} with the OneCycle learning rate scheduler~\cite{smith2019super} for 200 epochs. Following~\cite{simvp}, We choose the optimal learning rate from $\{1e^{-2}, 1e^{-3}, 1e^{-4}\}$ under the premise of stable training. The batch size is set to $16$ for all the models but $4$ for E3D-LSTM for its large memory cost. We evaluate the performance by mean square error (MSE), mean absolute error (MAE), and structural similarity index (SSIM)~\cite{wang2004image}. We repeat each experiment for three trials and report the average. FLOPs are reported using fvcore~\cite{wu2019detectron2}. The training time is reported by computing the average seconds for training an epoch and the inference efficiency is reported by inferencing 10,000 test samples with a batch size of 1 and computing the average testing frames per second (FPS). Experiments are conducted on a single NVIDIA V100 GPU.

\begin{figure*}[ht]
\centering
\includegraphics[width=0.98\textwidth]{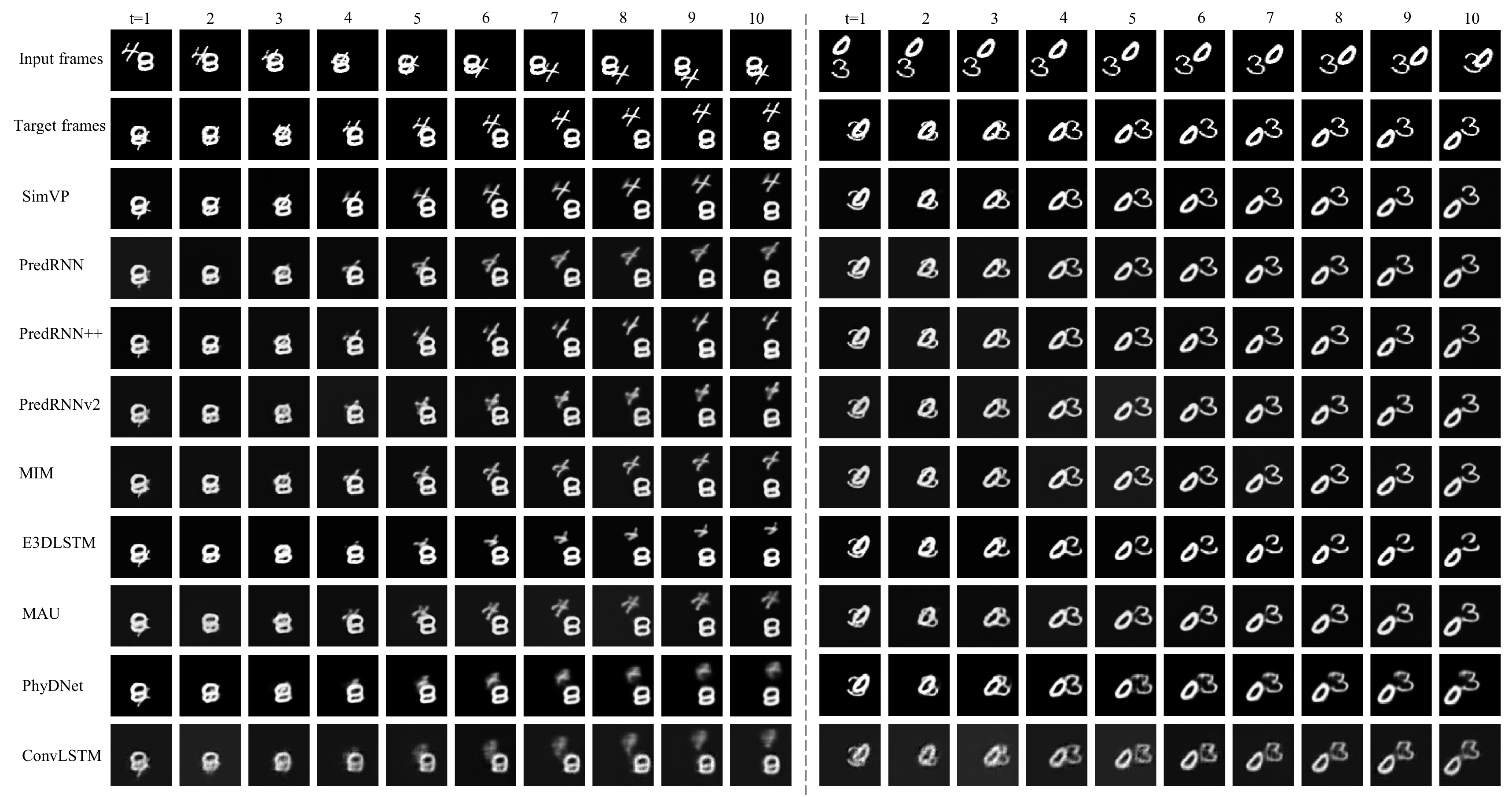} 
\caption{Examples of predicted results on the Moving MNIST dataset. We denote SimVPv2 as SimVP for convenience here.}
\vspace{-2mm}
\label{fig:example_mmnist} 
\end{figure*}

Table~\ref{tab:mmnist} shows the performance on the Moving MNIST dataset. SimVPs achieve highly competitive performance compared to state-of-the-art. For the small model group (the first five rows in Table~\ref{tab:mmnist}), SimVPv2-S obtains the best prediction quality with the fastest training time and the highest inference efficiency. For the large model group (from the sixth to the last row in Table~\ref{tab:mmnist}), SimVPv2-L, which uses larger hidden dimensions and a larger number of layers, achieves the best prediction quality compared with other oversized models. Furthermore, we report SimVPv2-S$\times 3$ that simply trains SimVPv2-S with three times epochs, i.e., 600 epochs. Surprisingly, SimVPv2-S$\times 3$ achieves competitive performance as well as SimVPv2-L. Morover, SimVPv2-S$\times 3$ still has the least training time compared with large models.

We plot the performance vs. training time and the performance vs. inference efficiency in Fig.~\ref{fig:train_inference_efficiency}. In Fig.~\ref{fig:train_inference_efficiency}(a), both the training time and MSE are the lower the better. We can see that SimVPs are concentrated in the lower-left corner of this plot. SimVPv2-S even takes about only one-sixteenth training time of E3D-LSTM and obtains significantly better performance. In Fig.~\ref{fig:train_inference_efficiency}(b), the inference efficiency is the higher the better. SimVPs are concentrated in the lower-right corner. SimVPs outperform other models and are the only model that can achieve more than 10 FPS. SimVPv2-S has about six times inference efficiency compared to ConvLSTM-S and about forty times compared to CrevNet. Based on the above observations, we demonstrate that SimVPs outperform state-of-the-art methods in both training and inference efficiency.

Fig.~\ref{fig:example_mmnist} shows the qualitative comparison between SimVP and other state-of-the-art methods. It can be seen that SimVP predicts much clearer frames, especially when it comes to long-range predictions. For the first example, only SimVP shows clear and sharp digit '4' while other methods do not. When digit '4' and digit '8' are overlapped at $t=4$, we still can infer these digits from the predicted frame of SimVP, but other methods fail to reconstruct the original digit '4'. PhyDNet and ConvLSTM even produce severely blurry frames from the beginning to the end. For the second example, most methods perform well except PhyDNet and ConvLSTM. PredRNN and its variants predict high-quality frames, but their predicted digits have some distortions, while SimVP keeps predicting almost the same frames as the ground-truth frames.

\begin{figure*}[ht]
\centering
\includegraphics[width=0.99\textwidth]{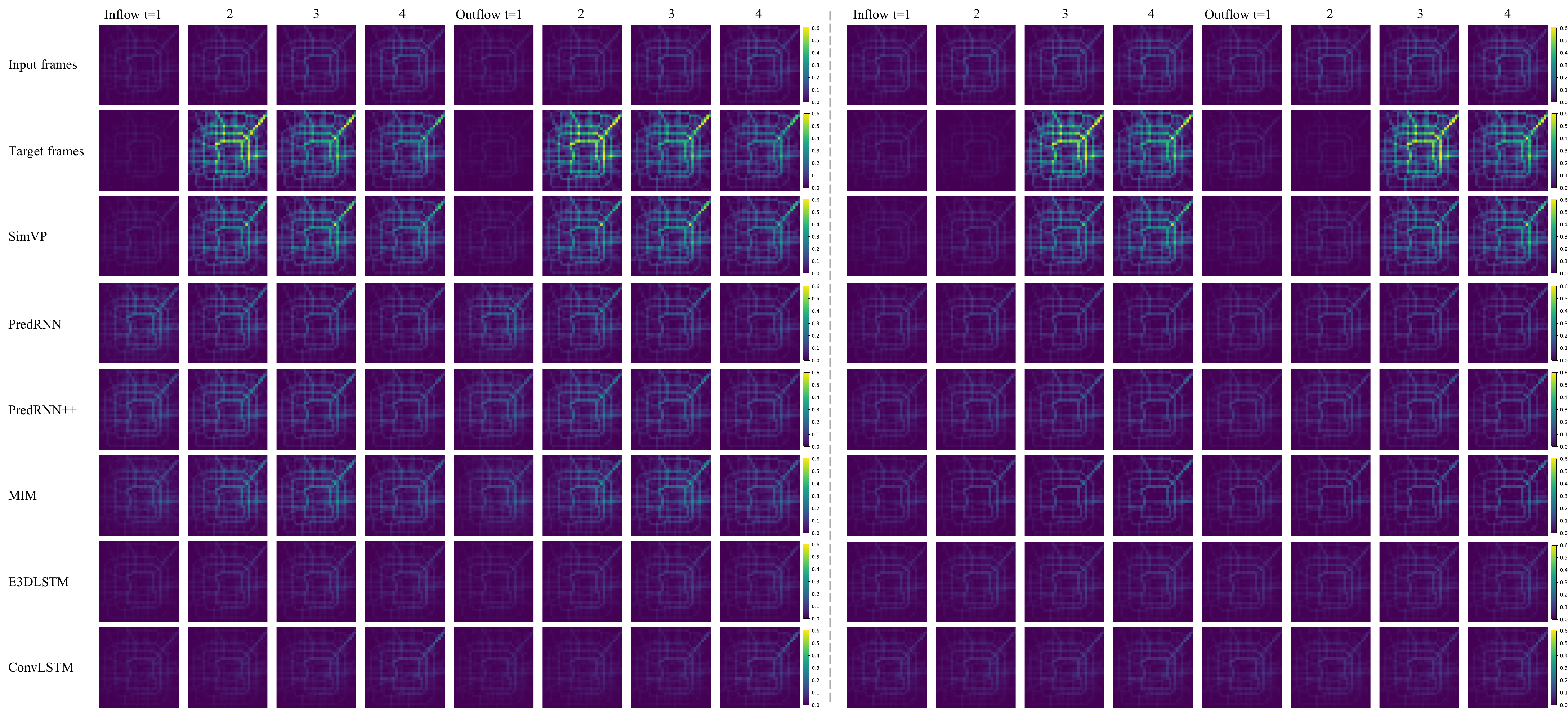}
\caption{Examples of predicted results on the TaxiBJ dataset. We denote SimVPv2 as SimVP for convenience here.}
\vspace{-2mm}
\label{fig:example_taxibj} 
\end{figure*}

\subsubsection{TaxiBJ}


We use the TaxiBJ dataset~\cite{zhang2017deep} to evaluate the traffic forecasting ability. TaxiBJ contains the trajectory data in Beijing collected from taxicab GPS with two channels, i.e., inflow or outflow defined in~\cite{zhang2017deep}. Models are trained to predict 4 subsequent frames by observing the prior 4 frames. We compare SimVP with ConvLSTM, PredRNN, PredRNN+, MIM, E3D-LSTM, and PhyDNet.

\begin{table}[ht]
\vspace{-2mm}
\centering
\caption{Quantitative results on the TaxiBJ dataset ($4 \rightarrow 4$ frames).}
\setlength{\tabcolsep}{5mm}{
\begin{tabular}{c ccc}
\toprule
& \multicolumn{3}{c}{TaxiBJ} \\
Method & MSE $\times$ 100$\downarrow$ & MAE$\downarrow$ & SSIM$\uparrow$ \\ 
\midrule
ConvLSTM   & 48.5  & 17.7 & 0.978 \\
PredRNN      & 46.4  & 17.1 & 0.971 \\
PredRNN++ & 44.8  & 16.9 & 0.977 \\
MIM              & 42.9  & 16.6 & 0.971 \\
E3D-LSTM     & 43.2  & 16.9 & 0.979 \\
PhyDNet      & 41.9  & 16.2 & 0.982 \\ 
\hline
SimVP & 41.4  & 16.2 & 0.982 \\
SimVPv2 & \textbf{34.8}  & \textbf{15.6} & \textbf{0.984} \\  
\bottomrule
\end{tabular}
}
\label{tab:traffic}
\vspace{-2mm}
\end{table}

As shown in Table~\ref{tab:traffic}, SimVP outperforms the previous recurrent-based state-of-the-art methods by a small margin. SimVPv2 which is introduced in this paper further stretches the margin between baseline models, quantitatively improving SimVP by about 15.94\% in the MSE metric and about 3.7\% in the MAE metric. Benefiting from the gated spatiotemporal attention mechanism, SimVPv2 is able to achieve superior performance on such a complex traffic flow forecasting problem.

We also visualize two examples of predicted results on the TaxiBJ in Fig.~\ref{fig:example_taxibj}. These two examples are exceptional cases that have very different target frames comparing to input frames. The first example has a sudden increase in traffic flow under both inflow and outflow channels from $t=2$ in the target frames. The second example also performs a similar trend as the first example, but from $t=3$ in the target frames. The predicted results of these two complex examples are impressive. While other recurrent-based methods fail to capture such different traffic variations, SimVP accurately predicts the future trend to a large extent and unexpectedly finds the sudden traffic jam from the observations of placid transportation. This phenomenon reveals the powerful perception of the long-range future of SimVP. SimVP learns the spatiotemporal dynamics in a way that is consistent with the real-world situation. In contrast, recurrent-based methods over-depend on the previous frames, and they are not able to directly capture the long-range dependencies in such complex traffic flows.

\subsubsection{WeatherBench}


We employ our model in the climate prediction on the WeatherBench~\cite{rasp2020weatherbench}. This dataset contains various types of climatic data from 1979 to 2018. The raw data is regrid to low resolutions, we here choose $5.625^{\circ}$ ($32 \times 64$ grid points) resolution for our data. We choose the temperature prediction task to evaluate our model. Following the protocol from~\cite{rasp2020weatherbench}, we train the model using data from 1979 to 2015 and validate the model using data from 2016. The evaluation is done for the years 2017 and 2018. We use the global temperature from the past 12 hours to predict that in the future 12 hours. The unit of global temperature is $K$. The results are evaluated by RMSE and MAE metrics. We compare our model with other strong climate prediction baselines, i.e., TGCN~\cite{tgcn}, STGCN~\cite{stgcn}, MSTGCN~\cite{guo2019attention}, ASTGCN~\cite{guo2019attention}, GCGRU~\cite{seo2018structured}, DCRNN~\cite{li2018diffusion}, AGCRN~\cite{bai2020adaptive}, CLCSTN~\cite{lin2022conditional}, and CLCRN~\cite{lin2022conditional}. Additional comparisons with spatiotemporal predictive learning methods such as ConvLSTM~\cite{convlstm}, PredRNN~\cite{predrnn}, and PredRNN++~\cite{predrnn++} are included. 

\begin{table}[h]
\centering
\caption{Quantitative results on the WeatherBench dataset ($12 \rightarrow 12$ frames).}
\setlength{\tabcolsep}{9mm}{
\begin{tabular}{c cc}
\toprule
& \multicolumn{2}{c}{WeatherBench} \\
Method & MAE$\downarrow$ & RMSE$\downarrow$ \\ 
\midrule
Copying   & 1.6906 & 2.4838 \\
TGCN   & 3.8638 & 5.8554 \\
STGCN  & 4.3525 & 6.8600 \\ 
MSTCN  & 1.2199 & 1.9203 \\
ASTGCN & 1.4896 & 2.4622 \\ 
GCGRU  & 1.3256 & 2.1721 \\
DCRNN  & 1.3232 & 2.1874 \\
AGCRN  & 1.2551 & 1.9314 \\
CLCSTN & 1.3325 & 2.1239 \\
CLCRN  & 1.1688 & 1.8825 \\
ConvLSTM  & 1.0529 & 1.4606 \\
PredRNN   & 0.8268 & 1.2119 \\
PredRNN++ & 0.8054 & 1.1776 \\
\hline
SimVP   & 0.7882 & 1.1483 \\
SimVPv2 & \textbf{0.7475}  & \textbf{1.0785} \\  
\bottomrule
\end{tabular}}
\label{tab:weather}
\end{table}


We report the quantitative results in Table~\ref{tab:weather}. As the input and target frames are similar, we also report the results by copying the input frames as the predicted frames to evaluate the actual results, which is denoted as 'Copying' in Table~\ref{tab:weather}. It can be seen that SimVP outperforms baselines by relatively large margins. Specifically, SimVP improves the state-of-the-art meteorological forecasting model CLCRN by about 36.04\% in the MAE metric and about 42.71\% in the RMSE. 


\subsection{Generalization ability across different datasets}
\label{lab:generalization}

Generalizing the knowledge across different datasets, especially in an unsupervised setting, is the core research point of machine learning and artificial intelligence. To investigate the generalization ability of SimVP, we train the model for 50 epochs on KITTI and evaluate it on Caltech Pedestrian. Both KITTI and Caltech datasets are captured from road traffic scenarios but in different environments. 

\begin{table}[h]
\centering
\caption{Quantitative results on the Caltech Pedestrian dataset ($10 \rightarrow 1$ frame).}
\setlength{\tabcolsep}{6mm}{
\begin{tabular}{c cccc}
\toprule
& \multicolumn{3}{c}{Caltech Pedestrian} \\
Method & SSIM$\uparrow$ & PSNR$\uparrow$ & LPIPS$\downarrow$ \\ 
\midrule
BeyondMSE  & 0.847 & - & - \\
MCnet      & 0.879 & - & - \\
DVF        & 0.897 & 26.2 & 5.57 \\
Dual-GAN   & 0.899 & - & - \\
CtrlGen    & 0.900 & 26.5 & 6.38 \\
PredNet    & 0.905 & 27.6 & 7.47 \\
ContextVP  & 0.921 & 28.7 & 6.03 \\
SDC-Net    & 0.918 & - & - \\
rCycleGan  & 0.919 & 29.2 & - \\
DPG        & 0.923 & 28.2 & 5.04 \\
CrevNet    & 0.925 & 29.3 & - \\
STMFANet   & 0.927 & 29.1 & 5.89 \\
\hline
SimVP      & 0.940 & 33.1 & 3.81 \\
SimVPv2    & \textbf{0.949} & \textbf{33.2} & \textbf{3.11} \\  
\bottomrule
\end{tabular}
}
\label{tab:caltech}
\end{table}

\begin{figure*}[ht]
\centering
\includegraphics[width=0.98\textwidth]{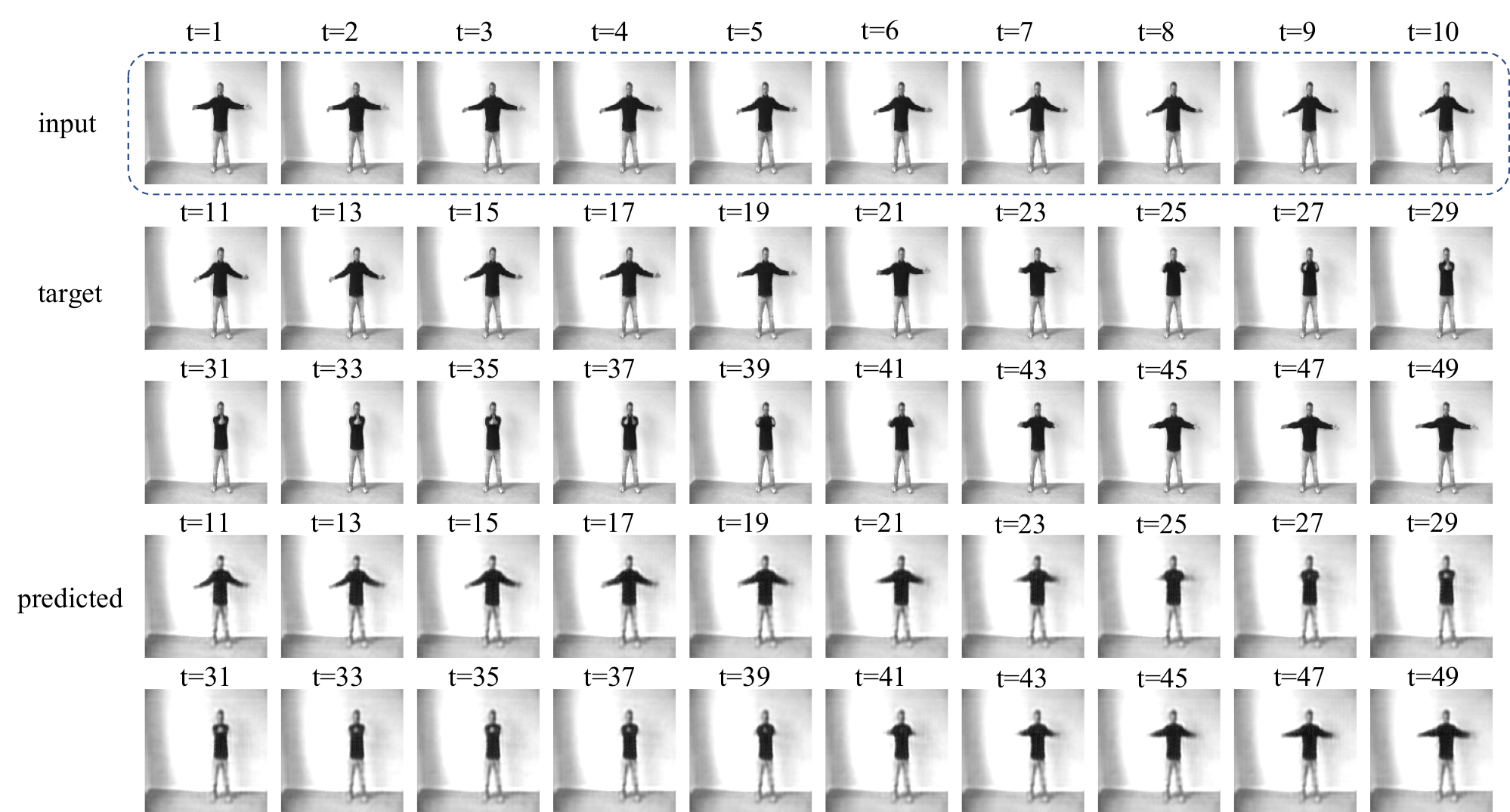}
\caption{An example of predicted results on the KTH dataset.}
\label{fig:example_kth} 
\end{figure*}


Following~\cite{lotter2016deep,crevnet,oprea2020review}, several strong baselines are selected for comparison, including BeyondMSE~\cite{mathieu2015deep}, MCnet~\cite{villegas2017decomposing}, DVF~\cite{liu2017video}, Dual-GAN~\cite{liang2017dual}, CtrlGen~\cite{hao2018controllable}, PredNet~\cite{lotter2016deep}, ContextVP~\cite{byeon2018contextvp}, SDC-Net~\cite{reda2018sdc}, rCycleGan~\cite{kwon2019predicting}, DPG~\cite{gao2019disentangling}, CrevNet~\cite{crevnet} and STMFANet~\cite{jin2020exploring}. SSIM~\cite{wang2004image}, PSNR, and LPIPS~\cite{zhang2018unreasonable} metrics are used in the evaluation phase. As shown in Table~\ref{tab:caltech}, SimVP has achieved better performance than baseline models by a large margin. Specifically, SimVP outperforms STMFANet by about 1.04\% in the SSIM, approximately 13.74\% in the PSNR, and about 35.31\% in the LPIPS. SimVPv2 further improves the SimVP and obtains the best performance among SSIM, PSNR, and LPIPS.


 
\subsection{Predicting frames with flexible lengths}
\label{lab:flexible}


We choose the KTH dataset~\cite{schuldt2004recognizing} for evaluating the flexibility. It contains 25 individuals performing 6 types of actions, i.e., walking, jogging, running, boxing, hand waving, and hand clapping. Following~\cite{e3dlstm,oprea2020review}, we compare the PSNR and SSIM of SimVP with other baselines on KTH. We train our model for 100 epochs and evaluate the results by SSIM and PSNR metrics. Models are trained to predict the next 20 or 40 frames from the previous 10 observations. Strong baselines are included, such as MCnet~\cite{villegas2017decomposing}, ConvLSTM~\cite{convlstm}, SAVP, SAVP-VAE~\cite{lee2018stochastic}, VPN~\cite{kalchbrenner2017video}, DFN~\cite{jia2016dynamic}, fRNN~\cite{oliu2018folded}, Znet~\cite{zhang2019z}, SV2P~\cite{babaeizadeh2017stochastic}, PredRNN~\cite{predrnn}, VarNet~\cite{jin2018varnet}, PredRNN++~\cite{predrnn++}, MSNET~\cite{lee2018mutual}, E3d-LSTM~\cite{e3dlstm}, and STMFANet~\cite{jin2020exploring}. We compare the predicted quality with these state-of-the-art baselines under both $10 \rightarrow 20$ frames and $10 \rightarrow 40$ frames cases.

\begin{table}[h]
\centering
\caption{Quantitative results on the KTH dataset ($10 \rightarrow 20/40$ frames).}
\setlength{\tabcolsep}{3.5mm}{
\begin{tabular}{ccccc}
\toprule
& \multicolumn{2}{c}{KTH ($10 \rightarrow 20$)} & \multicolumn{2}{c}{KTH ($10 \rightarrow 40$)} \\ \cline{2-5} 
Method  & SSIM$\uparrow$ & PSNR$\uparrow$ & SSIM$\uparrow$ & PSNR$\uparrow$ \\
\midrule
MCnet & 0.804 & 25.95 & 0.73 & 23.89 \\
ConvLSTM & 0.712 & 23.58 & 0.639 & 22.85 \\
SAVP & 0.746 & 25.38 & 0.701 & 23.97 \\
VPN & 0.746 & 23.76 & --  & --  \\
DFN & 0.794 & 27.26 & 0.652 & 23.01 \\
fRNN & 0.771 & 26.12 & 0.678 & 23.77 \\
Znet & 0.817 & 27.58 & --    & --    \\
SV2Pv & 0.838 & 27.79 & 0.789 & 26.12 \\
PredRNN     & 0.839 & 27.55 & 0.703 & 24.16 \\
VarNet & 0.843 & 28.48 & 0.739 & 25.37 \\
SAVP-VAE & 0.852 & 27.77 & 0.811 & 26.18 \\
PredRNN++ & 0.865 & 28.47 & 0.741 & 25.21 \\
MSNET & 0.876 & 27.08 & --    & --    \\
E3d-LSTM   & 0.879 & 29.31 & 0.810 & 27.24 \\
STMFANet & 0.893 & 29.85 & 0.851 & 27.56 \\ 
\midrule
SimVP & 0.905 & 33.72 & 0.886 & 32.93 \\
SimVPv2 & \textbf{0.913} & \textbf{34.24} & \textbf{0.895} & \textbf{33.35} \\
\bottomrule
\end{tabular}}
\label{tab:kth}
\end{table}

We show the quantitative results in Table~\ref{tab:kth}. It can be seen that SimVPs are superior to those baselines. Moreover, SimVPs even accurately predict the future frames under the extremely long-range case like $10 \rightarrow 40$ frames. We show the qualitative results on the KTH dataset with output frame lengths of 40 in Fig.~\ref{fig:example_kth}. In general, SimVP can predict the overall posture in almost every frame, albeit with a slight blur. The performance comes to the recursive prediction strategy that mitigates the issue of error accumulation. In typical recurrent models, predictions are generated frame by frame, leading to sequential error accumulation at each time step, which can severely degrade the quality of long-term forecasts. In contrast, our recursive approach reduces the frequency of error propagation by accumulating errors only over four iterations, as opposed to forty iterations in recurrent methods when predicting 40 frames ahead. Furthermore, gSTA modules dynamically adjusts its attention weights based on the evolving spatiotemporal patterns, enabling the model to adapt to changes in motion dynamics throughout the prediction sequence. Additionally, SimVP's architecture facilitates the seamless integration of spatiotemporal context, allowing the model to jointly learn spatial and temporal features without treating them as separate components.


\subsection{Challenging multi-domain evaluation}
\label{lab:challenging}

To thoroughly assess the robustness and adaptability of models, we conduct evaluations on two challenging multi-domain datasets: RoboNet and BridgeData. These datasets are specifically chosen due to their diverse range of tasks and environments, which present significant challenges. Both datasets involve predicting the next 10 frames given the first 2 input frames. We compare SimVPv2 with ConvLSTM, E3D-LSTM, MAU, PhyDNet, PredRNN, PredRNN++, and PredRNNv2 using SSIM and PSNR metrics. Table~\ref{tab:robonet} presents the quantitative results for the RoboNet and BridgeData datasets. The results demonstrate that SimVPv2 consistently outperforms other baseline models across both datasets, achieving the highest SSIM and PSNR values.

On the RoboNet dataset, SimVPv2 achieves an SSIM of 0.856 and a PSNR of 22.78, surpassing the performance of recurrent-based models such as PredRNN++ (SSIM 0.849, PSNR 22.66) and PredRNNv2 (SSIM 0.847, PSNR 22.52). On the BridgeData dataset, SimVPv2 attains an SSIM of 0.865 and a PSNR of 22.62, outperforming all other methods. Notably, it outperforms PredRNN++ (SSIM 0.856, PSNR 22.34) and ConvLSTM (SSIM 0.832, PSNR 21.36), indicating that the model can effectively generalize across a variety of domains and tasks with different visual characteristics. The superior performance of SimVPv2 on the challenging multi-domain evaluation illustrates the versatility and robustness of SimVPv2, establishing it as a strong baseline for future research in spatiotemporal predictive learning. The results suggest that even in complex, multi-domain settings, a streamlined model like SimVPv2 can outperform more complex architectures, reinforcing the benefits of simplicity and efficiency.

\begin{table}[h]
\centering
\caption{Quantitative results on the RoboNet ($2 \rightarrow 10$ frames) and BridgeData datasets ($2 \rightarrow 10$ frames).}
\setlength{\tabcolsep}{3.5mm}{
\begin{tabular}{ccccc}
\toprule
& \multicolumn{2}{c}{RoboNet ($2 \rightarrow 10$)} & \multicolumn{2}{c}{BridgeData ($2 \rightarrow 10$)} \\ \cline{2-5} 
Method  & SSIM$\uparrow$ & PSNR$\uparrow$ & SSIM$\uparrow$ & PSNR$\uparrow$ \\
\midrule
ConvLSTM  & 0.836 & 22.15 & 0.832 & 21.36 \\
E3D-LSTM  & 0.827 & 21.82 & 0.784 & 20.36 \\
MAU       & 0.843 & 22.40 & 0.821 & 21.14  \\
PhyDNet   & 0.797 & 20.92 & 0.762 & 19.61 \\
PredRNN   & 0.850 & 22.63 & 0.853 & 22.19 \\
PredRNN++ & 0.849 & 22.66 & 0.856 & 22.34  \\
PredRNNv2 & 0.847 & 22.52 & 0.850 & 22.01 \\
\midrule
SimVP   & 0.854 & 22.73 & 0.863 & 22.60 \\
SimVPv2 & \textbf{0.856} & \textbf{22.78} & \textbf{0.865} & \textbf{22.62} \\
\bottomrule
\end{tabular}}
\label{tab:robonet}
\end{table}

\subsection{Ablation study}

\subsubsection{The quantitative analysis of spatiotemporal translator}

The flexibility of our framework allows for the seamless integration of various spatiotemporal translators, enabling a comprehensive comparison of different architectures. We explore the adaptability of our model by replacing the spatiotemporal translator with different architectures, including vanilla ViT~\cite{dosovitskiy2020image}, Swin Transformer~\cite{liu2021swin}, Poolformer~\cite{yu2022metaformer}, MLPMixer~\cite{tolstikhin2021mlp}, and ConvMixer~\cite{trockman2022patches}. This analysis helps to evaluate the impact of spatiotemporal translator choices. The results indicate that the gSTA module consistently outperforms other architectures across all evaluation metrics, achieving the lowest MSE and MAE values while also obtaining the highest SSIM score. Notably, gSTA achieves these results with comparable FLOPs and parameter count relative to other competitive architectures, such as Swin Transformer and MLPMixer. This suggests that gSTA effectively captures spatiotemporal dependencies with greater accuracy and efficiency. In comparison, ConvMixer shows the lowest FLOPs and parameter, making it the most lightweight option; however, its predictive performance is lower than that of gSTA and some other architectures. Swin Transformer and MLPMixer also demonstrate strong predictive capabilities, with MLPMixer achieving an MSE close to gSTA but requiring more computational resources. IncepU, on the other hand, exhibits higher computational complexity and does not match the performance with gSTA.

Overall, these results underscore the effectiveness of the gSTA module in capturing complex spatiotemporal relationships while maintaining computational efficiency, making it a suitable choice for high-performance temporal modeling.

\begin{table}[h]
\centering
\caption{Ablation study on the Moving MNIST dataset.}
\setlength{\tabcolsep}{1mm}{
\begin{tabular}{ccccccc}
\toprule
Method & Params (M) & FLOPs (G) $\downarrow$ & MSE $\downarrow$ & MAE $\downarrow$ & SSIM $\uparrow$ \\
\midrule
ViT       & 46.1 & 16.9 & 35.15 & 95.87 & 0.914 \\
Swin Transformer & 46.1 & 16.4 & 29.70 & 84.05 & 0.933 \\
Poolformer & 37.1 & 14.1 & 31.79 & 88.48 & 0.927 \\
MLPMixer  & 38.2 & 14.7 & 29.52 & 83.36 & 0.934 \\
ConvMixer & \textbf{3.9} & \textbf{5.5} & 32.09 & 88.93 & 0.926 \\
\midrule 
IncepU    & 58.0 & 19.4 & 32.22 & 89.19 & 0.927 \\
gSTA      & 46.8 & 16.5 & \textbf{26.60} & \textbf{77.32} & \textbf{0.940} \\
\bottomrule
\end{tabular}}
\label{tab:ablation}
\end{table}  

\subsubsection{The qualitative analysis of modules}

\begin{figure}[ht]
\centering
\includegraphics[width=0.34\textwidth]{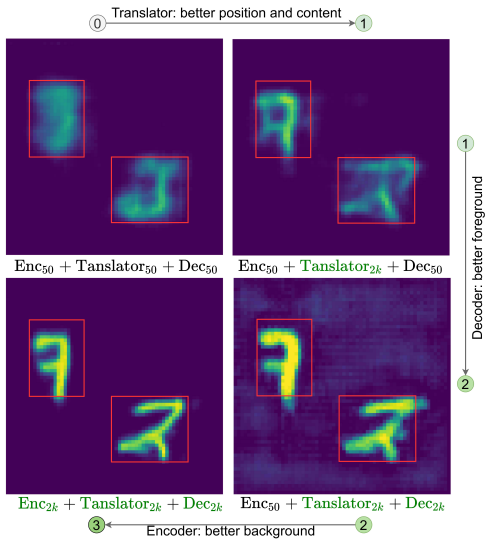}
\caption{The role of the Translator, Spatial Encoder and Decoder.}
\label{fig:ablation} 
\end{figure}

To explore the roles of spatiotemporal translator, spatial encoder, and decoder, we perform an ablation study on the Moving MNIST dataset, as shown in Fig.~\ref{fig:ablation}. We represent submodules trained with $n$ epochs as $\mathrm{Enc}_n, \mathrm{Translator}_n, \mathrm{Dec}_n$. Given a model trained with $50$ epochs, we replace its submodules with maturer ones trained with $2,000$ epochs. It can be seen that the spatiotemporal translator focuses on predicting the position and content of the objects. The spatial encoder focuses on the background portrayed, and the spatial decoder is responsible for optimizing the shape of the foreground objects.

\section{Conclusion}

In this paper, we introduce SimVPv2, aimed at pushing the boundaries of existing spatiotemporal predictive learning models. Building on the success of SimVP, which demonstrated the feasibility of using a fully convolutional architecture to enable parallel processing and reduce dependence on complex recurrent structures, SimVPv2 goes a step further by completely removing Unet-like multi-scale architectures. Instead, it incorporates an efficient gSTA mechanism, allowing SimVPv2 to achieve state-of-the-art performance with a more streamlined and computationally efficient design. Through extensive experiments on the synthetic moving digits, traffic flow forecasting, climate prediction, road driving, human motion prediction, and robo action planning, we demonstrate the superior performance of SimVPv2 under various settings like standard spatiotemporal predictive learning, generalization across similar scenarios, prediction with flexible lengths, and multi-domain evaluation. We believe SimVPv2 establishes a strong baseline that will benefit future research in spatiotemporal predictive learning.



%



\ifCLASSOPTIONcompsoc
  \section*{Acknowledgments}
\else
  \section*{Acknowledgment}
\fi

This work was supported by National Science and Technology Major Project (No. 2022ZD0115101), National Natural Science Foundation of China Project (No. 624B2115, No. U21A20427), Project (No. WU2022A009) from the Center of Synthetic Biology and Integrated Bioengineering of Westlake University and Integrated Bioengineering of Westlake University and Project (No. WU2023C019) from the Westlake University Industries of the Future Research Funding.

\ifCLASSOPTIONcaptionsoff
  \newpage
\fi



%
\bibliographystyle{IEEEtran}
\bibliography{ref}



%





\end{document}